\documentclass[10pt]{article}

\usepackage{PRIMEarxiv}

\usepackage[utf8]{inputenc} % allow utf-8 input
\usepackage[T1]{fontenc}    % use 8-bit T1 fonts
\usepackage{hyperref}       % hyperlinks
\usepackage{url}            % simple URL typesetting
\usepackage{booktabs}       % professional-quality tables
\usepackage{amsfonts}       % blackboard math symbols
\usepackage{nicefrac}       % compact symbols for 1/2, etc.
\usepackage{enumitem}
\usepackage{lipsum}
\usepackage{multirow}
\usepackage{lineno}
\usepackage{amsmath, amssymb, mathrsfs, bbm}
\usepackage{tabularx}
\usepackage{setspace}
\usepackage{subcaption}
\usepackage{xcolor}         % use colors
\usepackage{fancyhdr}       % header
\usepackage{graphicx}       % graphics
\usepackage[backend=biber,style=ieee,sorting=none,natbib=true]{biblatex}
\usepackage{authblk}
\usepackage{algorithm}
\usepackage{algpseudocode}
\usepackage{cleveref}
\usepackage{arydshln}

\Crefname{figure}{Fig.}{Figs.} 
\Crefname{table}{Table}{Tables}
\Crefname{algorithm}{Alg.}{Algs.}
\Crefname{equation}{Eq.}{Eqs.}

\graphicspath{{media/}}     % organize your images and other figures under media/ folder
\usepackage{bm}

\title{Cott-ADNet: Lightweight Real-Time Cotton Boll and Flower Detection Under Field Conditions}
\author[1]{Rui-Feng Wang}
\author[1]{Mingrui Xu}
\author[2]{Matthew C Bauer}
\author[2]{Iago Beffart Schardong}
\author[3]{Xiaowen Ma}
\author[4,*]{Kangning Cui}
\affil[1]{ Department of Crop and Soil Sciences, University of Georgia}
\affil[2]{ Institute of Plant Breeding, Genetics and Genomics, University of Georgia-Tifton Campus}
\affil[3]{ School of Software Technology, Zhejiang University}
\affil[4]{ Department of Computer Science, Wake Forest University}

\begin{document}
\maketitle

\let\thefootnote\relax
\noindent\footnotetext{$*$ Corresponding author: cuij@wfu.edu}

\begin{abstract}
Cotton is one of the most important natural fiber crops worldwide, yet harvesting remains limited by labor-intensive manual picking, low efficiency, and yield losses from missing the optimal harvest window. Accurate recognition of cotton bolls and their maturity is therefore essential for automation, yield estimation, and breeding research. We propose Cott-ADNet, a lightweight real-time detector tailored to cotton boll and flower recognition under complex field conditions. Building on YOLOv11n, Cott-ADNet enhances spatial representation and robustness through improved convolutional designs, while introducing two new modules: a NeLU-enhanced Global Attention Mechanism to better capture weak and low-contrast features, and a Dilated Receptive Field SPPF to expand receptive fields for more effective multi-scale context modeling at low computational cost. We curate a labeled dataset of 4,966 images, and release an external validation set of 1,216 field images to support future research.  Experiments show that Cott-ADNet achieves 91.5\% Precision, 89.8\% Recall, 93.3\% mAP50, 71.3\% mAP, and 90.6\% F1-Score with only 7.5 GFLOPs, maintaining stable performance under multi-scale and rotational variations. These results demonstrate Cott-ADNet as an accurate and efficient solution for in-field deployment, and thus provide a reliable basis for automated cotton harvesting and high-throughput phenotypic analysis. Code and dataset is available at \url{https://github.com/SweefongWong/Cott-ADNet}.
\end{abstract}

\noindent \textbf{Index Terms}: 
cotton, cotton boll detection, lightweight object detection, rotational convolution

\section{Introduction}
\label{sec:intro}
Cotton is one of the most critical economic crops worldwide, accounting for nearly 35\% of global natural fiber production. It underpins industries such as textiles, apparel, and healthcare, while also serving as a primary source of income for over 100 million farmers \cite{W1, 3}. However, the harvesting process remains a major bottleneck. Manual picking is labor-intensive, costly, and unsuitable for large-scale production \cite{7, 2}. Moreover, the timing of harvest is crucial for fiber quality, and manual operations often miss the optimal window, which leads to significant yield and quality losses \cite{8, W1}. These challenges highlight the urgent need for automated vision systems that can reliably identify cotton bolls and their maturity status to support harvesting, yield estimation, and breeding research.

Previous studies on cotton boll recognition have explored both traditional machine learning and deep learning approaches, yet major obstacles remain for practical deployment in the field \cite{W3, 11, V2, 16}. Existing detectors already provide fast and lightweight inference, and several studies have adapted them to agricultural tasks~\cite{V1, 17, 18, c1, c2}. However, real-world field conditions introduce persistent difficulties: (i) occlusion and dense distributions, where overlapping bolls are easily overlooked; (ii) illumination variability and background clutter, which degrade visual features; and (iii) lack of validation in real-world field settings, as most prior work relied mainly on curated datasets. These challenges are further compounded by the small size and irregular distribution of cotton bolls, which makes accurate detection particularly difficult \cite{24, 25}. Prior improvements such as lightweight compression or generic attention modules mitigate some issues, but often sacrifice robustness or leave weak-feature and multi-scale context challenges insufficiently addressed~\cite{c3, 14, V10, V11, V3}. This gap motivates the design of Cott-ADNet, which incorporates architectural refinements to better capture weak features and multi-scale context while retaining the lightweight efficiency needed for deployment on edge devices.

To address the aforementioned challenges, this study proposes a lightweight framework for cotton boll detection with the following key contributions:

\begin{enumerate}
    \item \textbf{Task-driven architectural refinement:} Enhanced convolutional designs improve the representation of small, densely distributed bolls with varying orientations.
    
    \item \textbf{New modules for feature enhancement: }We introduce a NeLU-enhanced Global Attention Mechanism (NGAM) to strengthen weak feature representation and stabilize training, together with a Dilated Receptive Field SPPF (DRFSPPF) to expand receptive fields for effective multi-scale context modeling.

    \item \textbf{Lightweight yet balanced performance:} Extensive experiments and ablation studies show that Cott-ADNet surpasses existing detectors by achieving the highest F1-score, while also reaching 93.3\% mAP50 with only 7.5 GFLOPs. This demonstrates that our model delivers both efficiency and accuracy.

    \item \textbf{Field validation:} Beyond a curated dataset of 4,966 labeled images, we additionally test on 1,216 field images and provide qualitative evidence of model performance under real deployment conditions.
\end{enumerate}

In summary, this study offers both technical improvements and a practical step toward deployable agricultural vision systems that support automated cotton harvesting, yield monitoring, and high-throughput phenotyping.

\section{Related Work}
\label{sec:related work}

%-------------------------------------------------------------------------
\subsection{Cotton Boll Detection}

Early attempts at cotton boll detection relied on traditional machine learning approaches such as color thresholding and random forest classifiers \cite{V5, V6, 11}. While these approaches enabled initial progress toward automated harvesting, they suffer from limited scalability, poor adaptability to varying field conditions, and an inability to satisfy real-time requirements on edge devices \cite{V2, W4, 13, 14, c4}.

More recently, deep learning has demonstrated significant advantages in cotton boll detection and emerged as the dominant paradigm for automated harvesting. Variants of the YOLO (You Only Look Once) family have been applied to cotton-specific tasks, including unopened boll detection with YOLOv5 \cite{V7}, boll-splitting stage recognition with YOLOv8 \cite{24}, and disease classification with YOLOv11 \cite{V8}. Although these works demonstrate the potential of end-to-end detectors, most still lack lightweight optimization for agricultural deployment and have not been validated under real-world field conditions.

%-------------------------------------------------------------------------
\subsection{Lightweight Object Detection}

With the advancement of precision agriculture, the deployment of unmanned ground vehicles (UGVs) and unmanned aerial vehicles (UAVs) in the field has imposed increased demand for lightweight agricultural object detection deep learning model designs \cite{V13, V14, V15}. The YOLO series has become the mainstream solution for agricultural visual recognition because of its compact design, fast inference, and strong real-time performance \cite{W7, c2}.

In recent years, extensive efforts have been made to further optimize YOLO models by reducing computational overhead and enhancing their deployability in field environments. For example, \citeauthor{V10} have compressed the YOLOv5s model and accelerated inference by redesigning its network architecture and loss functions \cite{V10}. \citeauthor{V11} have reduced parameter count and computational complexity by streamlining the YOLOv8 detection head \cite{V11}. Furthermore, improvements to the YOLOv8 backbone and head combined with pruning strategies have been shown to simultaneously decrease model size and inference speed \cite{V12}. These optimizations demonstrate the value of lightweight design, but they often trade off detection robustness in agricultural environments where small-object recognition and feature degradation remain unsolved issues.

%-------------------------------------------------------------------------
\subsection{Attention and Convolution Enhancements}

Attention mechanisms and convolutional innovations have been widely explored to improve agricultural object detection. Attention mechanisms improve the model’s ability to focus on salient regions and suppress background noise. Meanwhile, advanced convolutional designs improve feature expressiveness without dramatically increasing cost. For example, Omni-Dimensional Dynamic Convolution (ODConv) \cite{W19} and Space-to-Depth Convolution (SPDConv) \cite{W21} have been shown to enhance spatial representation while remaining efficient.

Recent studies have demonstrated that integrating SPDConv with a small-object detection head can significantly improve the accuracy of cotton bolls detection during the boll-splitting stage \cite{24}. Moreover, the incorporation of a custom-designed attention mechanism has been shown to enhance feature extraction ability while improving computational efficiency \cite{V12}. Despite these advances, most existing designs remain generic and are not tailored to the unique challenges of cotton boll imagery, namely weak visual features, irregular spatial patterns, and real-world deployment demands. These gaps motivate Cott-ADNet, which introduces specialized attention and convolutional refinements to strengthen weak-feature expressiveness, expand receptive fields, and maintain lightweight efficiency.

\section{Methodology}
\label{methodology}

\begin{figure*}
    \centering
    \includegraphics[width=1.0\linewidth]{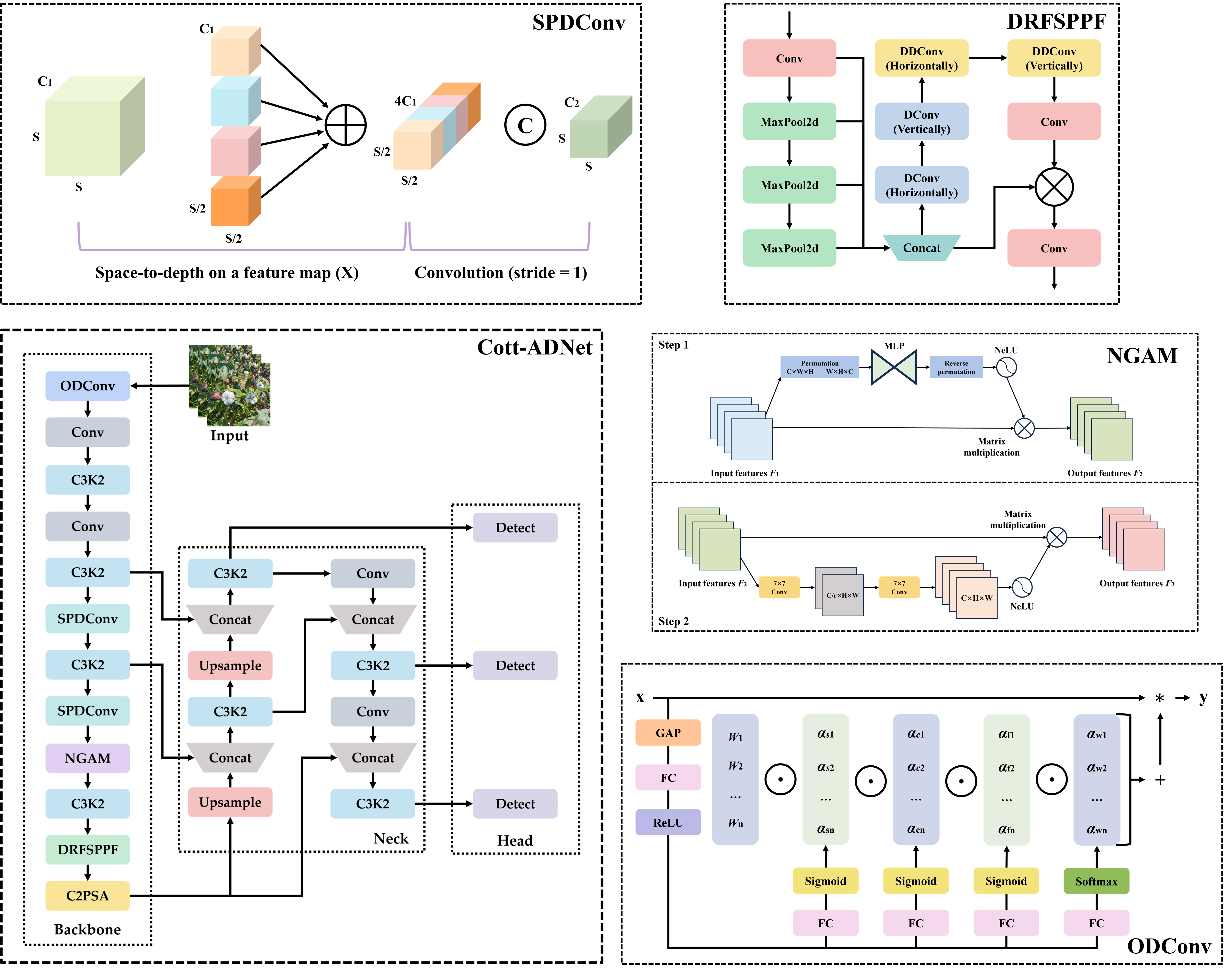}
    \caption{Overview of the proposed Cott-ADNet framework and its key modules.}
    \label{fig:main}
\end{figure*}

\subsection{NeLU-enhanced Global Attention Mechanism}

In complex scenarios such as cotton boll detection, effective modeling requires a balance between global semantic understanding and fine-grained feature representation. To address this, we integrated the Global Attention Mechanism (GAM) \cite{W27} into the deep layers of the YOLOv11n backbone. GAM consists of two complementary branches: a channel attention branch that captures inter-channel dependencies, and a spatial attention branch that encodes spatial contextual information. However, the original GAM relies on Sigmoid or ReLU activations, which suffer from gradient saturation and the “dying ReLU” problem that limits feature extraction. To mitigate these issues, this study incorporated the Negative-slope Linear Unit (NeLU) \cite{V9}, a non-linear function that preserve gradient flow for negative inputs. Its definition and derivative are given in \Cref{eq:NeLU} and \Cref{eq:dxNeLU}.

\begin{equation}\label{eq:NeLU}
    \text{NeLU}(x) = 
    \begin{cases} 
    x, & x > 0 \\
    -\dfrac{\alpha}{1 + x^2}, & x \leq 0
\end{cases}
\end{equation}

\begin{equation}\label{eq:dxNeLU}
    \frac{d}{dx} \text{NeLU}(x) = 
    \begin{cases} 
    1, & x > 0 \\
    \frac{2\alpha x}{(1 + x^2)^2}, & x \leq 0
\end{cases}
\end{equation}

In NeLU, the negative slope coefficient $\alpha$ was set to 0.2 based on empirical validation~\cite{V9}. For $x>0$. NeLU behaves identically to ReLU; for $x<0$, it provides a continuous non-zero negative gradient, allowing neurons to remain active during training. Compared to Sigmoid, NeLU avoids gradient saturation caused by exponential operations and reduces the associated computational overhead, offering improved numerical stability and inference efficiency.

Building upon this, we proposed NGAM (NeLU-enhanced Global Attention Mechanism), where all the activation functions within both the channel and spatial branches of the original GAM were uniformly replaced with NeLU. This modification (1) enhanced the representation of weak-textured and low-contrast objects such as cotton bolls and flowers; (2) improved the smoothness of deep feature interactions by the attention mechanism; and (3) stabilized the training process by ensuring non-zero gradients for negative inputs. These improvements come with minimal computational overhead. As a result, NGAM is well-suited for deployment on resource-constrained platforms such as UGVs and UAVs. Overall, NGAM preserved the global modeling capability of GAM while further strengthening feature expressiveness and cross-domain generalization through the integration of NeLU. Its structure is illustrated in \Cref{fig:main}.

\subsection{Dilated Receptive Field SPPF}

\begin{algorithm}[t]
\caption{Dilated Receptive Field SPPF (DRFSPPF)}
\label{alg:drfsppf}
\begin{algorithmic}[1]
\Require Feature map $F^C \in \mathbb{R}^{H \times W \times C}$
\Ensure Enhanced feature representation $\text{DRFSPPF}(F^C)$

\State \textbf{Multi-scale pooling:} Apply pooling with kernels $\{5,9,13\}$ and concatenate:
\Statex \centerline{$F^{C'} = \text{Concat}(F^{C}, P_{5}(F^C), P_{9}(F^C), P_{13}(F^C))$}

\State \textbf{Large-kernel DConv:} Apply depthwise conv ($k=11$):
\Statex \centerline{$F_{\text{DConv}}^C=W_{(2d-1)\times1}^C \ast (W_{1 \times (2d-1)}^C \ast F^{C'})$}

\State \textbf{Dilated DConv:} Expand receptive field with dilation $d$:
\Statex \centerline{$F_{\text{DDConv}}=W_{[11/d] \times 1}^C \ast (W_{1 \times [11/d]}^C \ast F_{\text{DConv}}^C)$}

\State \textbf{Residual recalibration:} Fuse with $F^C$:
\Statex \centerline{$\bar{F}^C=(W_{1\times1} \ast F_{\text{DDConv}}) \odot F^C$}

\State \textbf{Output projection:} Final representation:
\Statex \centerline{$\text{DRFSPPF}(F^C) = W_{1 \times 1}^{out} \ast \bar{F}^C$}

\end{algorithmic}
\end{algorithm}

Accurate cotton boll detection requires not only identifying small, low-contrast targets but also capturing broader contextual structures such as leaves and stems. The original Spatial Pyramid Pooling–Fast (SPPF) module \cite{V27, V28} provides efficient multi-scale feature aggregation, but its receptive field remains limited. To address this, we extended SPPF into a Dilated Receptive Field SPPF (DRFSPPF), which combined large-kernel and dilated convolutions to enlarge the effective receptive field while maintaining lightweight efficiency.

As outlined in \Cref{alg:drfsppf}, DRFSPPF first applied multi-scale pooling with kernels $\{5,9,13\}$ and concatenated the resulting features with the original input. A large-kernel depthwise convolution (DConv) with kernel size $k=11$ was subsequently applied to model long-range dependencies along both horizontal and vertical directions. This was followed by a dilated depthwise convolution (DDConv) with dilation rate $d$, which further expanded the receptive field and enhanced cross-region information aggregation. A residual recalibration was subsequently introduced via a Hadamard product with the original feature map, and finally, a $1\times1$ convolution produced the channel-wise output representation. The resulting feature maps combined enriched global semantics with preserved local details.

Here, $F^C$ is the input feature map of channel $C$; $P_{k_i}(\cdot)$ denotes pooling with kernel size $k_i$; $W$ are learnable convolution kernels; $\odot$ indicates element-wise multiplication; $F^{C'}$ denotes the feature map after pooling and concatenation; $F_{\text{DConv}}^C$ and $F_{\text{DDConv}}$ are the outputs of large-kernel and dilated depthwise convolutions, respectively; $\bar{F}^C$ is the recalibrated feature map; and $\ast$ represents convolution.

This design provided three major benefits. First, the combination of directional separable and depthwise separable convolutions substantially reduced the parameter count and FLOPs associated with large-kernel operations, which ensures efficiency. Second, the incorporation of dilated convolutions expanded receptive fields without introducing additional parameters, while the residual recalibration stabilized gradient flow during training. Third, the enlarged receptive field improved the joint modeling of small-scale targets and large-scale structures, which is essential in cluttered field environments. As a result, DRFSPPF is particularly well-suited for real-time detection tasks on resource-constrained platforms such as UAVs and field-deployed devices, while maintaining robustness under conditions typical of cotton fields.

\subsection{Proposed Model (Cott-ADNet)}

To address the challenges of small object size, subtle inter-class differences, and complex background interference in cotton boll and flower recognition, we proposed Cott-ADNet: a lightweight and high-precision detection model built on YOLOv11n \cite{V16}. As shown in \Cref{fig:main}, Cott-ADNet introduced targeted architectural refinements, including dynamic convolution, spatially reconstructive convolution, attention mechanisms, and enhanced global modeling, to improve robustness under complex field conditions.

At the backbone level, Cott-ADNet retained the overall architecture of YOLOv11n while introducing targeted modifications to address different semantic hierarchies. Specifically, ODConv \cite{W19} was integrated in the shallow feature extraction stage. By introducing learnable weights across spatial dimensions, input channels, output channels, and kernel dimensions, ODConv enabled joint modeling of multi-dimensional information and improved adaptability to directional structural variations (as shown in \Cref{eq:ODConv}).
\begin{equation}\label{eq:ODConv}
    \mathrm{ODConv}(X) = \sum_{i=1}^{K} \alpha_i \cdot \left( W_i \ast X \right)
\end{equation}
where $X$ is the input feature map, $W_i$ is the $i^{th}$ learnable convolution kernel; $\alpha_{i} \in \mathbb{R}$ is the dynamic weight generated by a directional-attention mechanism, $\ast$ denotes standard convolution, and $K$ is the number of kernels.

During the mid-to-high-level semantic modeling stages, Cott-ADNet incorporated SPDConv at two downsampling locations to mitigate the loss of small objects during compression \cite{W21}. By remapping spatial dimensions into the channel dimension prior to convolution, SPDConv achieved lossless downsampling. Specifically, the SPD operation firstly partitioned the input feature map $X \in \mathbb{R}^{S \times S \times C}$ into non-overlapping sub-blocks according to a downsampling factor $scale$, and reassembled them along the channel dimension to produce a new feature map $X'$ (as shown in \Cref{eq:SPDConv1}). Subsequently, a standard stride-1 convolution was applied to $X'$ for compressive encoding (as illustrated in \Cref{eq:SPDConv2}).

\begin{equation}\label{eq:SPDConv1}
    X' = \mathrm{SPD}(X; scale) \in \mathbb{R}^{\frac{S}{scale} \times \frac{S}{scale} \times C \cdot scale^2}
\end{equation}

\begin{equation}\label{eq:SPDConv2}
    X'' = \mathrm{Conv}_{1\times 1,\, C_{\mathrm{out}}} \left( X' \right) \in \mathbb{R}^{\frac{S}{\text{scale}} \times \frac{S}{\text{scale}} \times C_{\mathrm{out}}}
\end{equation}

In the deep feature fusion stage, Cott-ADNet incorporated NGAM to capture global contextual information and enhance feature responses to blurred or low-contrast targets from complex backgrounds. Simultaneously, the original SPPF module was replaced with DRFSPPF, which combined multi-scale pooling with large-kernel directional modeling to significantly expand the receptive field and strengthen global semantic representation.

In summary, Cott-ADNet maintained controllable parameter size and inference speed while effectively enhancing spatial representation and feature aggregation. This was achieved through directional modeling with ODConv, lossless downsampling with SPDConv, global semantic enhancement with NGAM, and long-range dependency modeling with DRFSPPF. These improvements provided a robust visual perception foundation for accurate recognition of cotton bolls and flowers, supporting subsequent automated harvesting and phenotypic analysis.

\section{Numerical Experiments}
\label{sec:results}

\subsection{Datasets and Field Test}
\label{data-field}

\begin{figure}[t]
  \centering
  % \fbox{\rule{0pt}{2in} \rule{0.9\linewidth}{0pt}}
   \includegraphics[width=0.5\linewidth]{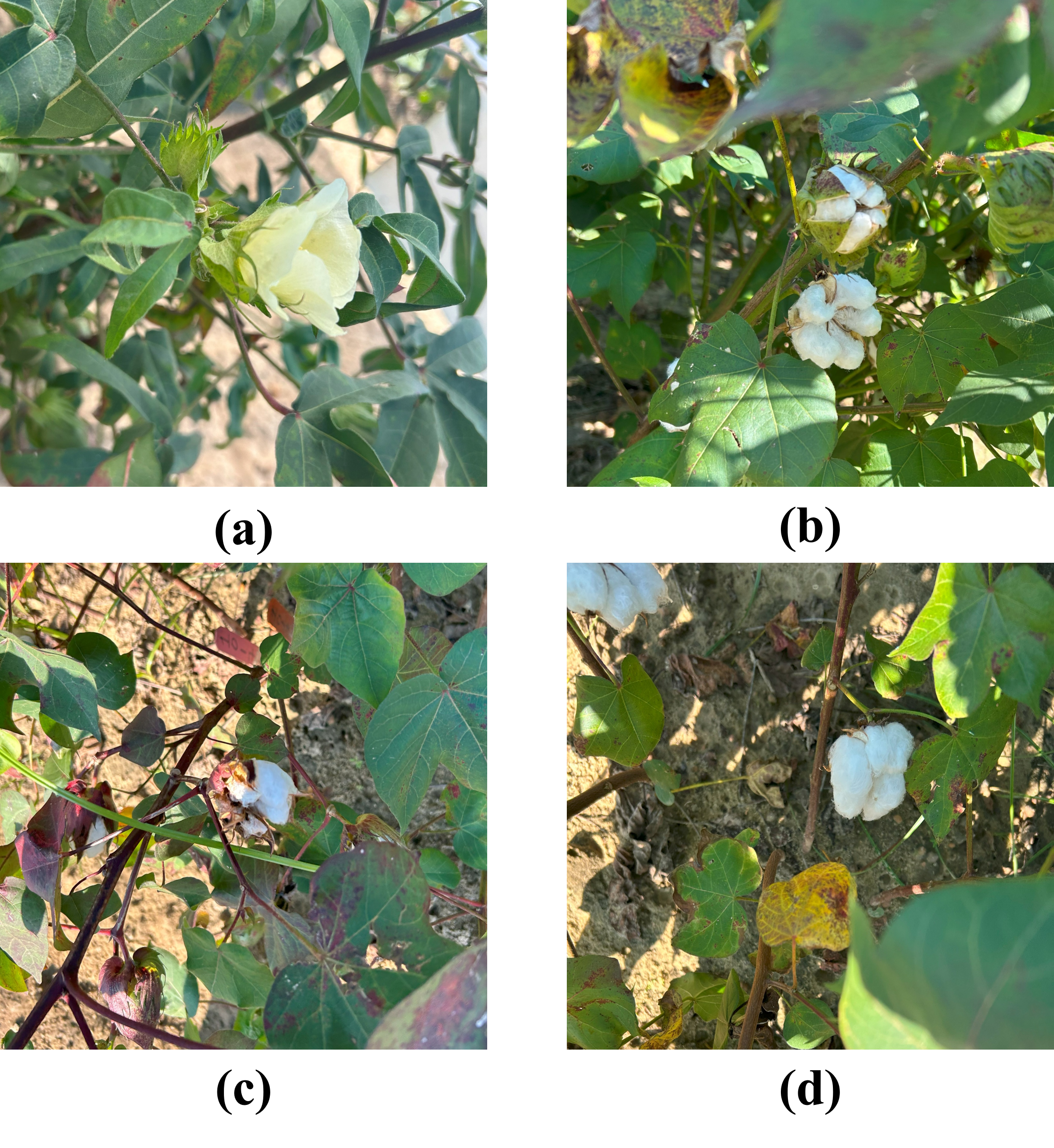}
   \caption{Samples of independent external test dataset: (a) FL — Cotton flower; (b) PB — Partly opened cotton boll; (c) DB — Defected cotton boll; (d) FB — Fully opened cotton boll.}
   \label{fig:sample}
\end{figure}

% The proposed model was trained and validated on our constructed cotton bolls and flowers recognition dataset of 4,966 images (3,982 for training and 984 for validation). In addition, we collected an independent external test dataset of 1,216 images from the University of Georgia-Tifton Campus Gibbs Research Farm located in Tifton City, Tift County, Georgia State 31793, United States of America (31$^\circ$26$'$13$''$ N, 83$^\circ$35$'$18$''$ W) (\Cref{fig:sample}). Furthermore, this external test dataset has been made publicly available to facilitate future research.

% the University of Georgia Gibbs Research Farm in Tifton, Georgia, USA (31$^\circ$26$'$13$''$ N, 83$^\circ$35$'$18$''$ W) using an iPhone 14 camera (3024 $\times$ 4032 or 4032 $\times$ 3024 pixels). 

To support both model development and practical evaluation, we constructed two complementary datasets. The dataset covered four target categories: cotton flowers (FL), partly opened bolls (PB), defected bolls (DB), and fully opened bolls (FB). These classes capture the key stages and conditions of cotton boll development relevant for automated harvesting. The first dataset served as the primary training resource, while the second was designed as an independent external set to assess generalization in real-world field conditions.

The primary dataset was built from two sources: publicly available repositories and web-scraped imagery. First, we collected open-source cotton boll detection datasets from platforms such as GitHub and Kaggle, and performed thorough cleaning to remove low-quality, mislabeled, or duplicate samples. Second, a Python-based crawler was developed to automatically acquire cotton-related images from multiple online sources, followed by manual inspection to ensure both quality and diversity. To further enhance annotation reliability, all labels were cross-validated by domain experts with experience in cotton breeding. Images were standardized to a resolution of 640 $\times$ 640 pixels in \texttt{jpg} format, and annotations were converted into the YOLO format~\cite{V28, c0}. To improve robustness, web-scraped images were augmented using random brightness adjustment, flipping, random masking, and Gaussian noise injection. The final dataset comprised 4,966 images, of which approximately 78\% originated from open-source repositories and 22\% from web-scraped sources. These were partitioned into training (3,982), validation (494), and test (490) subsets in an 8:1:1 ratio.

In addition, we collected an independent external dataset of 1,216 images at the University of Georgia Gibbs Research Farm in Tifton, Georgia, USA (31$^\circ$26$'$13$''$ N, 83$^\circ$35$'$18$''$ W). Images were captured with a handheld mobile camera under diverse field conditions, including variable lighting and occlusion. Metadata such as exact location and device model are withheld to preserve double-blind review. This dataset was used exclusively for evaluation and not involved in model development, thereby providing a more realistic measure of cross-domain generalization. Representative samples are shown in \Cref{fig:sample}.

\begin{table*}[htbp]
\centering
\caption{Comprehensive Evaluation of Cott-ADNet: Benchmark Comparisons and Ablation Study}
\label{tab:eval_comparison}
\begin{tabular}{lcccccc}
\hline
\textbf{Model} & $\boldsymbol{P}$ (\%) & $\boldsymbol{R}$ (\%) & $\boldsymbol{F_1}$ (\%) &
\textbf{mAP50 (\%)} & \textbf{mAP (\%)} & \textbf{GFLOPs} \\
\hline
YOLOv8s           & 90.271  & \textbf{89.516}  & 89.892  & 92.053  & 72.214  & 28.8  \\
YOLOv9s           & 89.997  & 89.176  & 89.585  & 91.917  & 72.983  & 27.6  \\
YOLOv10s          & 92.230  & 87.371  & 89.790  & 92.581  & 72.451  & 24.8  \\
YOLOv11s          & \textbf{92.615}  & 88.204  & \textbf{90.395}  & \textbf{93.417}  & \textbf{73.658}  & 21.6  \\
YOLOv12s          & 92.320  & 87.596  & 89.945  & 92.014  & 73.374  & 21.5  \\
RT-DETR-50        & 91.166  & 88.333  & 89.727  & 91.672  & 71.805  & 130.5 \\
\hdashline
YOLOv8n           & \textbf{93.577}  & 86.693  & 90.022  & 92.308  & 71.674  & 8.9   \\
YOLOv9t           & 90.313  & 88.109  & 89.197  & 92.317  & \textbf{73.637}  & 8.5   \\
YOLOv10n          & 90.026  & \textbf{90.579}  & 90.289  & 92.416  & 72.825  & 8.4   \\
YOLOv11n          & 90.236  & 88.224  & 89.227  & 92.197  & 71.844  & 6.4   \\
YOLOv12n          & 92.654  & 88.285  & \textbf{90.466}  & \textbf{92.795}  & 72.640  & \textbf{6.5}   \\
\hline
Remove ODConv     & 92.334  & 87.097  & 89.639  & 92.571  & 72.527  & 7.6   \\
Remove SPDConv    & \textbf{93.433}  & 87.970  & 90.619  & 92.199  & 71.202  & 7.9   \\
Remove DRFSPPF    & 90.486  & 86.211  & 88.297  & 90.925  & 71.068  & 7.3   \\
Remove NGAM       & 90.232  & 87.399  & 88.793  & 92.045  & 72.068  & \textbf{6.2}   \\
Remove NeLU       & 91.736  & 88.885  & 90.288  & 93.077  & \textbf{72.664}  & 7.5   \\
\textbf{Cott-ADNet}      & 91.543  & \textbf{89.753}  & \textbf{90.639}  & \textbf{93.285}  & 71.296  & 7.5   \\
\hline
\end{tabular}
\end{table*}

\subsection{Experiment Setup}

Experiments were conducted on a single NVIDIA V100 GPU, with data preprocessing and augmentation parallelized on an 8-core CPU. Training was performed with a batch size of 32 for up to 800 epochs. We used the AdamW optimizer with an initial learning rate of 0.001, weight decay of 0.0005, and L2 regularization. Early stopping with a patience of 50 was applied to prevent overfitting. All models were trained under the same pipeline to ensure reproducibility and fair comparison.

\subsection{Evaluation Metrics}

To evaluate the effectiveness of Cott-ADNet in cotton boll detection, we adopted standard object detection metrics, including Precision ($P$), Recall ($R$), F1-score ($F_1$), mean Average Precision (mAP), and computational complexity in terms of GFLOPs. Specifically, mAP was reported as mAP50 (IoU = 0.5) and as the average over IoU thresholds from 0.5 to 0.95 with a step of 0.05. The metrics were defined as follows:  
\[
P = \frac{tp}{tp + fp}, \quad 
R = \frac{tp}{tp + fn}, \quad 
F_1 = \frac{2PR}{P+R},
\]
where $tp$, $fp$, and $fn$ denote true positives, false positives, and false negatives, respectively. FLOPs denote the total number of floating-point operations required for a single forward pass at the input resolution, normalized as GFLOPs for efficiency comparison of computational complexity.

\subsection{Comparative Experiments}

To evaluate the effectiveness and efficiency of the proposed Cott-ADNet model in cotton bolls and flowers detection, we compared it against a range of contemporary object detection models, including the YOLOv8/9/10/11/12 series (both n- and s-variants) \cite{V17, V18, V19, V16, V20, c0, V28}, as well as the Transformer-based RT-DETR-50 \cite{V21}. As summarized in \Cref{tab:eval_comparison}, the models can be naturally grouped into three categories: (1) small-size detectors above the dashed line, which achieve strong accuracy but incur higher computational cost (typically $>$20 GFLOPs, with RT-DETR-50 exceeding 130 GFLOPs); (2) lightweight tiny-size baselines between the dashed and solid line, where FLOPs are below 10 and thus more suitable for edge deployment; and (3) ablation variants below the solid line, used to isolate the contribution of each proposed module.

Within this context, Cott-ADNet achieves the highest $F_1$ (90.639\%) among all models and the highest mAP50 (93.285\%) among detectors with fewer than 10 GFLOPs. This indicates that Cott-ADNet maintains a well-balanced trade-off between precision and recall, which enables it to recover more true positives without sacrificing stability. Compared with larger models above the dashed line, the gap in mAP50 is negligible: for example, YOLOv11s achieves the best score (93.417\%), only 0.13\% higher than Cott-ADNet, but requires nearly three times the FLOPs (21.6 vs. 7.5). Such differences are practically insignificant given the computational advantage of Cott-ADNet in resource-constrained deployment scenarios.

When compared against other lightweight baselines ($<$10 GFLOPs), Cott-ADNet surpasses all alternatives in both $F_1$ and mAP50. Although YOLOv8n yields the highest precision (93.577\%) and YOLOv10n achieves the highest recall (90.579\%), both models fail to balance these metrics, leading to lower $F_1$ than Cott-ADNet. Similarly, YOLOv9t shows $\sim$1\% higher mAP (73.637\%) under stricter IoU thresholds, but this comes at the expense of lower $F_1$ and mAP50, suggesting weaker robustness in practical detection. In contrast, Cott-ADNet again provides a more consistent and balanced performance profile, which makes it particularly suitable for real-time field deployment.

\begin{figure*}[t]
    \centering
    \includegraphics[width=\linewidth]{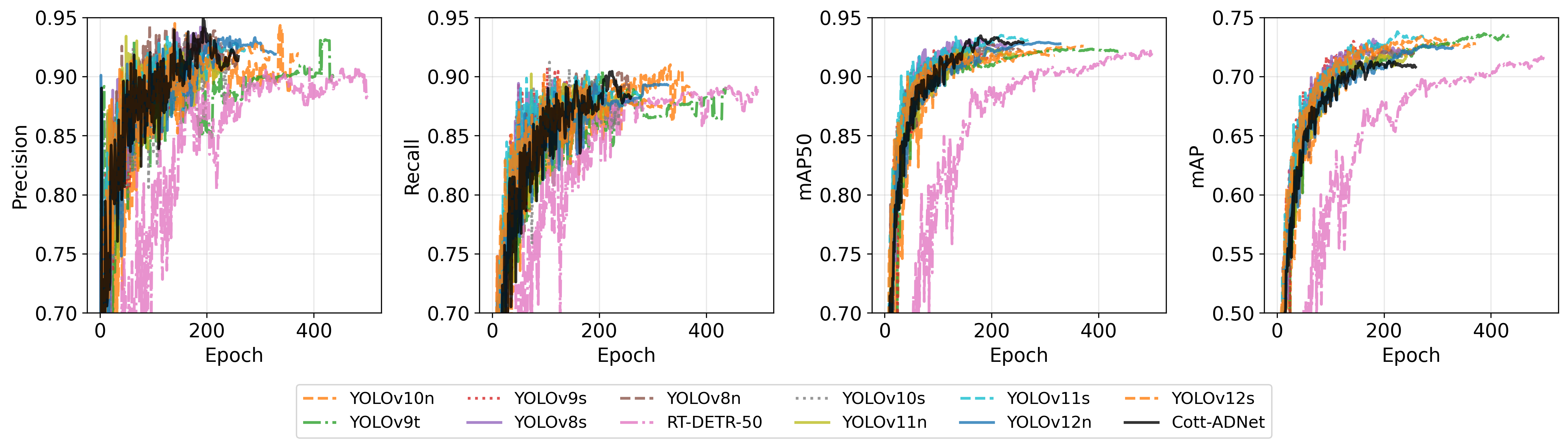}
    \caption{Validation performance comparison of different detection models during training. The curves show the evolution of Precision, Recall, mAP50, and mAP over epochs.}
    \label{fig:validation_metrics}
\end{figure*}

In addition, \Cref{fig:validation_metrics} illustrates the training dynamics of all models. Cott-ADNet converged reliably within 259 epochs, which is on par with or faster than most baselines. In contrast, YOLOv9t required 437 epochs (about 1.7$\times$ longer) to converge despite its higher mAP. YOLOv12n, although reporting lower GFLOPs, needed 328 epochs while still producing lower mAP50. Relative to YOLOv11n, Cott-ADNet trained only 15 additional epochs but improved nearly all metrics by over 1\% at the small cost of 1.1 extra GFLOPs. Compared to YOLOv10n, Cott-ADNet achieved +1.0\% higher mAP50 while also reducing computational cost by 0.9 GFLOPs. Notably, even lightweight baselines such as YOLOv8n, YOLOv9t and YOLOv10n consumed more GFLOPs than Cott-ADNet. These convergence results further demonstrate that Cott-ADNet provides a favorable balance of accuracy, efficiency, and training stability.

\begin{figure*}[t]
    \centering
    \includegraphics[width=\textwidth]{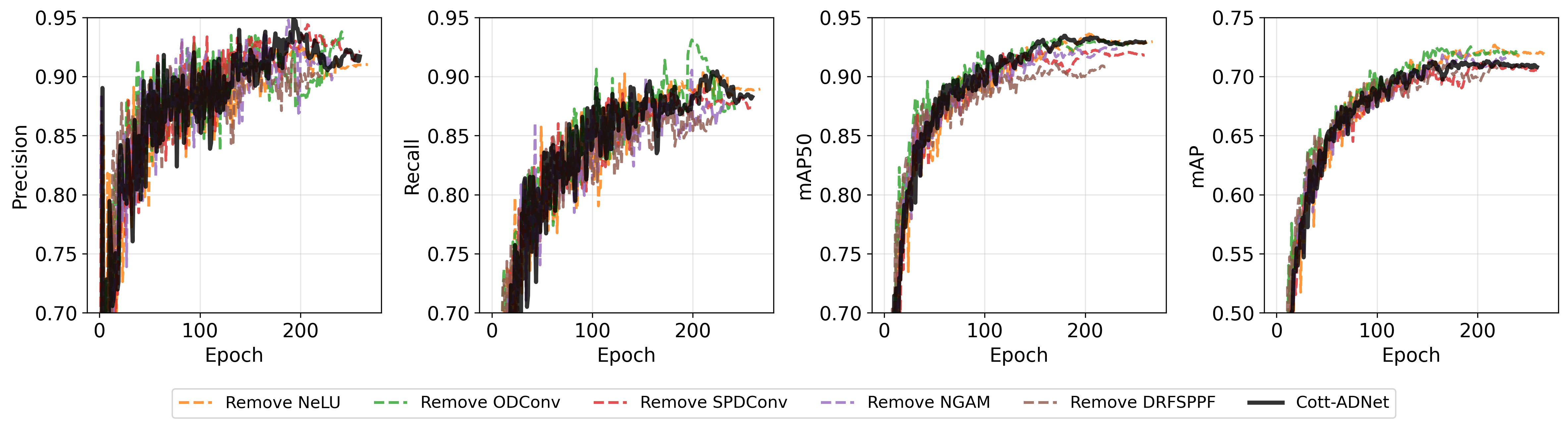}
    \caption{Ablation study results on validation metrics. Curves show the evolution of precision, recall, mAP50, and mAP across epochs for Cott-ADNet and its ablated variants.}
    \label{fig:ablation_metrics}
\end{figure*}

\subsection{Ablation Experiments}

\begin{figure*}
    \centering
    \includegraphics[width=1.0\linewidth]{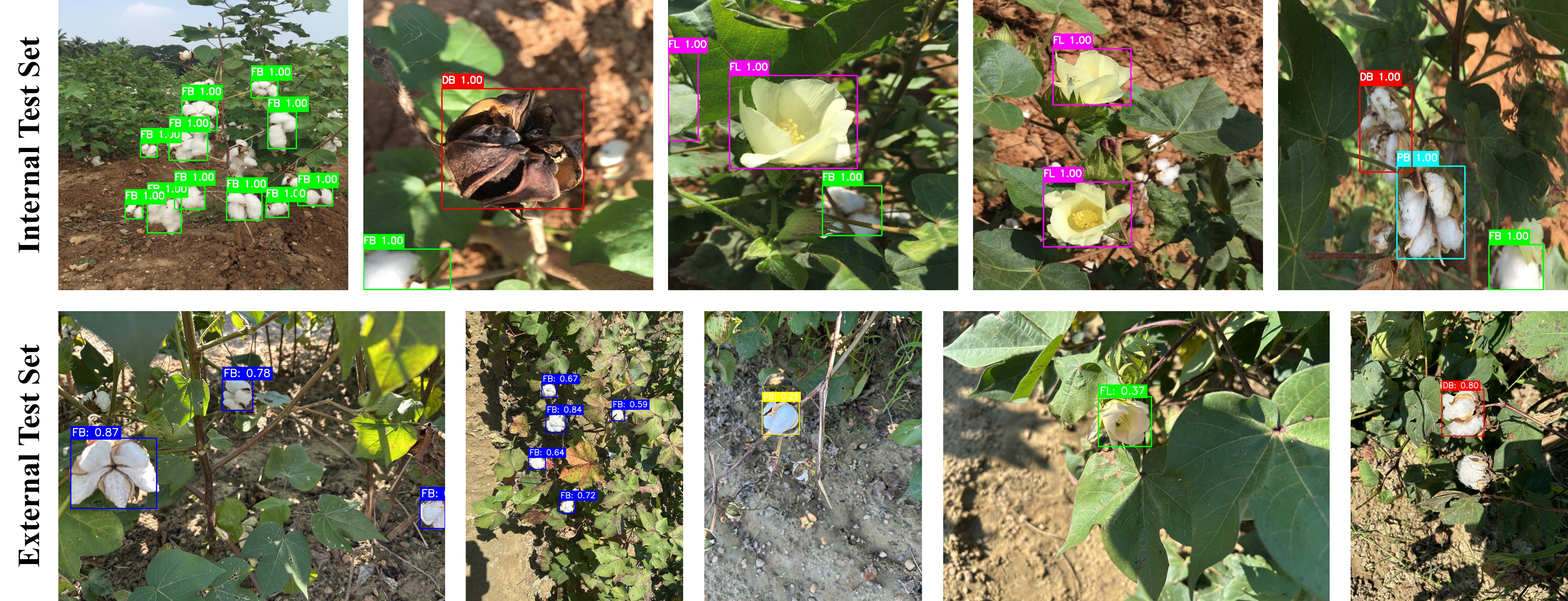}
    \caption{Detection results of Cott-ADNet on the Internal Test Set and External Test Set.}
    \label{fig:external}
\end{figure*}

To validate the effectiveness of each key component in Cott-ADNet, we performed ablation experiments in which ODConv, SPDConv, NGAM, NeLU, and DRFSPPF were individually removed and re-trained under identical settings. As reported in the lower part of \Cref{tab:eval_comparison}, ODConv and SPDConv clearly contributed to lightweight efficiency. Removing either module increased GFLOPs while also slightly lowering overall performance. This shows that the proposed lightweight convolutions not only reduce computational cost but also preserve accuracy. In contrast, DRFSPPF increased GFLOPs slightly (+0.2) yet provided one of the largest boosts in mAP50 (+2.36\% vs.\ the ablated variant) and $F_1$ (+2.34\%), effectively compensating for the minor trade-off from lightweight convolutions. This shows its efficacy on multi-scale feature aggregation. Similarly, NGAM was the heaviest component (reducing FLOPs to 6.2 when removed), but its absence led to substantial accuracy drops across all metrics (–1–2\%), which confirms its strong contribution to robust feature modeling. NeLU, although lightweight, improved training balance: its removal yielded higher mAP but less stable precision–recall trade-offs, whereas its inclusion kept the model more consistent without adding complexity.

Convergence behavior further supports these findings. As shown in \Cref{fig:ablation_metrics}, Cott-ADNet converged in a similar number of epochs to the base model YOLOv11n, while ablated variants such as those without NGAM (232 epochs) and without DRFSPPF (220 epochs) reached convergence slightly faster but at the cost of lower accuracy. The mAP50 curve highlights Cott-ADNet’s lead over all ablated models, confirming the importance of DRFSPPF and NGAM. Their slight training overhead is justified by consistent gains in robustness and overall performance.

\subsection{External Test}

We further evaluated the generalization of Cott-ADNet by comparing performance on the Internal Test Set and the independently collected External Field Test Set (See \Cref{data-field}). As illustrated in \Cref{fig:external}, results on the internal set were highly confident and accurate, reflecting the relatively controlled and less complex conditions of these images. On the external set, despite more complex backgrounds and smaller target sizes, the model maintained strong performance when objects were clearly visible, producing reliable localization and high-confidence predictions.

Performance degradation was mainly observed in the most challenging cases. Severe occlusion by foliage or branches occasionally led to missed detections of flowers or bolls, while illumination artifacts such as reflections and shadows reduced confidence and caused misclassifications (e.g., confusion between partly opened and fully opened bolls). These findings indicate that Cott-ADNet demonstrates robust cross-domain transfer but still faces challenges in highly heterogeneous environments.

\section{Conclusions}
\label{conclusions}
We presented Cott-ADNet, a lightweight and high-precision detector for cotton boll and flower recognition under complex field conditions. By integrating lightweight convolution modules with the proposed NGAM and DRFSPPF, the model achieves a strong balance between accuracy and efficiency. Experiments demonstrated that Cott-ADNet consistently outperforms lightweight YOLO baselines in $F_1$ and mAP50 while maintaining only 7.5 GFLOPs. This efficiency ensures that the model is suitable for real-time deployment on agricultural edge devices. These results highlight the potential of Cott-ADNet to support automated harvesting, yield monitoring, and high-throughput phenotyping in practical cotton production.

Future work will expand the scale and diversity of field data through the construction of a publicly available benchmark dataset with expert annotated labels that captures more complex real-world conditions. In addition, we plan to deploy and evaluate Cott-ADNet on UAV and UGV platforms for in-field testing. Beyond these steps, future directions include integrating multi-modal data (e.g., thermal imagery) to improve robustness under illumination variation, investigating domain adaptation techniques for cross-region generalization, and extending the framework to end-to-end yield estimation pipelines.

\newpage

% -------------------------------------------------------------------------

\printbibliography 

\end{document}